\PassOptionsToPackage{square,numbers,sort&compress}{natbib}
\documentclass[final, 3p,times,12pt,authoryear]{elsarticle}

\usepackage{amssymb}
\usepackage{amsmath}
\usepackage{caption}
\usepackage{booktabs}
\usepackage{longtable}
\usepackage{multirow}
\usepackage{array}
\usepackage{graphicx}
\usepackage{subcaption}
\usepackage{float}
\usepackage{hyperref}
\usepackage[utf8]{inputenc}
\usepackage[T1]{fontenc}
\usepackage{times}
\usepackage[square,numbers,sort&compress]{natbib}
\setcitestyle{numbers,square,comma}
\usepackage{pdfpages}
\usepackage[table,xcdraw]{xcolor}
\journal{Accident Analysis and Prevention}
\begin{document}
\begin{frontmatter}

\title{ROAR: Robust Accident Recognition and Anticipation for Autonomous Driving}

\author{Xingcheng Liu\textsuperscript{a}, Yanchen Guan\textsuperscript{a}, Haicheng Liao\textsuperscript{a}, Zhengbing He\textsuperscript{b}, Zhenning Li\textsuperscript{*,c}} 

\affiliation{organization={State Key Laboratory of Internet of Things for Smart City and Department of Computer and Information Science, University of Macau},
            city={Macau SAR},
            country={China}}

\affiliation{organization={Senseable City Lab, Massachusetts Institute of Technology}, 
            city={Cambridge MA}, 
            country={United States}}

\affiliation{organization={State Key Laboratory of Internet of Things for Smart City and Departments of Civil and Environmental Engineering and Computer and Information Science, University of Macau},
            city={Macau SAR},
            country={China}}

\begin{abstract}
Accurate accident anticipation is essential for enhancing the safety of autonomous vehicles (AVs). However, existing methods often assume ideal conditions, overlooking challenges such as sensor failures, environmental disturbances, and data imperfections, which can significantly degrade prediction accuracy. Additionally, previous models have not adequately addressed the considerable variability in driver behavior and accident rates across different vehicle types. To overcome these limitations, this study introduces \textbf{ROAR}, a novel approach for accident detection and prediction. ROAR combines Discrete Wavelet Transform (DWT), a self-adaptive object-aware module, and dynamic focal loss to tackle these challenges. The DWT effectively extracts features from noisy and incomplete data, while the object-aware module improves accident prediction by focusing on high-risk vehicles and modeling the spatial-temporal relationships among traffic agents. Moreover, dynamic focal loss mitigates the impact of class imbalance between positive and negative samples. Evaluated on three widely used datasets—Dashcam Accident Dataset (DAD), Car Crash Dataset (CCD), and AnAn Accident Detection (A3D)—our model consistently outperforms existing baselines in key metrics such as Average Precision (AP) and mean Time-to-Accident (mTTA). These results demonstrate the model’s robustness in real-world conditions, particularly in handling sensor degradation, environmental noise, and imbalanced data distributions. This work offers a promising solution for reliable and accurate accident anticipation in complex traffic environments.
\end{abstract}

\begin{highlights}
\item Application of Discrete Wavelet Transform for robust feature extraction from noisy and incomplete data.
\item Dynamic focal loss for addressing class imbalance, emphasizing hard-to-classify instances.
\item Model robustness in challenging scenarios, including sensor degradation, noise, and imbalanced datasets.
\item Superior performance on real-world datasets, surpassing baselines in key metrics.
\end{highlights}


\end{frontmatter}

\section{INTRODUCTION}
\label{sec1}

Traffic accidents are a persistent global issue, causing significant harm to both individuals and society. With the rise of autonomous driving, the need to proactively address this challenge has never been more pressing \cite{abdel2024matched}. Accident anticipation — the ability to predict vehicular collisions before they occur — has emerged as a cornerstone of autonomous vehicle safety systems, providing a mechanism to prevent accidents rather than merely detect them after the fact \cite{ref208,ref209}. By forecasting potential hazards, autonomous vehicles can take preventative actions that significantly reduce the risk of collisions, making traffic systems safer for all road users \cite{ref211}. Yet, achieving this goal is a complex and multifaceted problem, demanding sophisticated models capable of handling the unpredictable dynamics of real-world traffic environments.
The inherent difficulty in anticipating accidents lies in the dynamic nature of traffic flow, compounded by diverse driver behaviors, rapidly changing road conditions, and unpredictable interactions between vehicles, pedestrians, and environmental elements \cite{ref213,ref215}. While substantial progress has been made in accident detection, the ability to predict accidents ahead of time is still fraught with challenges \cite{wang2024deepaccident}. These include the limitations of existing sensor technologies, the dynamic interaction between road users, and the challenge of detecting subtle cues that often precede accidents. For instance, many accidents are caused by seemingly minor factors, such as a slight swerve, a delay in response to a traffic signal, or the sudden appearance of an obstacle. Traditional systems often fail to capture these nuanced events, leading to delays in intervention that could otherwise prevent an accident from occurring \cite{ref5,ref23}.

\begin{figure}
\centering
\includegraphics[width=1\textwidth]{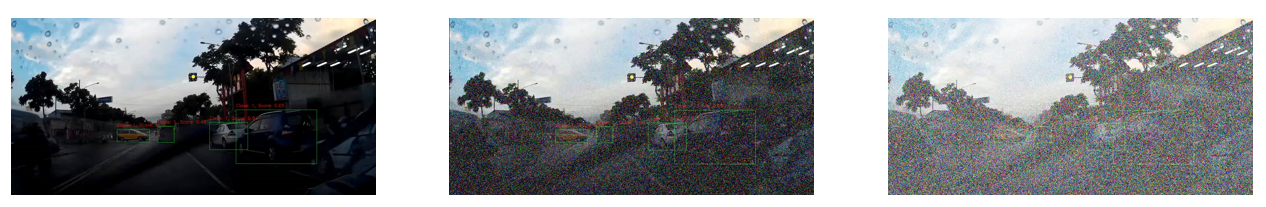}
\caption{Comparison of different levels of sensor distortion in the same scene relative to the raw video. The quality of the signal transmitted by the sensor directly impacts the input data, which in turn affects the model's ability to accurately anticipate accidents.}
\label{tab:Fig. 0}
\end{figure}
Moreover, many existing accident detection datasets are collected under highly controlled conditions, where sensor performance is ideal, and environmental factors such as weather or road obstructions are minimal \cite{fang2021dada}. In real-world driving scenarios, however, sensor degradation (for instance, Fig. \ref{tab:Fig. 0}), malfunctions, and the impact of environmental noise such as rain, fog, or road construction can significantly degrade the quality of the data, making accurate prediction even more challenging. Current models are trained and validated under these idealized conditions, leading to overoptimistic performance evaluations that fail to account for real-world variability. This discrepancy between ideal training conditions and real-world application highlights the need for robust models that can adapt to noisy, incomplete, and unreliable data.

Class imbalance further complicates accident prediction, particularly in datasets with high accident rates. For instance, in some classical dataset such as A3D (AnAn Accident Detection) \cite{ref1}, where the accident rate exceeds 90\%, the overwhelming presence of accident instances (positive samples) can lead to models that predict accidents for almost every instance, effectively overfitting to the dominant class \cite{fang2022traffic}. This high accident rate introduces a bias in the model’s learning process, where the model becomes overly sensitive to predicting accidents, often at the cost of accuracy and generalizability. Such a bias results in models that may predict accidents for most instances, even when they are not likely to occur, reducing their ability to identify non-accident scenarios that are crucial for precise and timely intervention \cite{karim2023attention}. Consequently, while the model might achieve a high overall accuracy by predicting most events as accidents, this approach leads to poor model performance when distinguishing between genuine accidents and non-events. Addressing this imbalance is crucial for developing models that not only predict accidents reliably but also avoid false positives and make more timely and accurate predictions across both frequent and rare events.

To effectively tackle these intertwined challenges, we propose a novel framework that enhances the accuracy, robustness, and adaptability of accident anticipation models. At the heart of our approach lies the Discrete Wavelet Transform (DWT), a powerful tool for extracting multi-resolution features from traffic data. By applying DWT, we enable our model to capture both high-frequency subtle differences between features (such as discriminative details of different frames) and low-frequency global pattern of feature vectors (such as category dominant features), making the model more resilient to sensor errors and environmental noise. DWT’s ability to break down signals into different frequency components provides a more comprehensive understanding of the data, allowing our model to maintain performance even in challenging conditions where traditional methods may fail.

However, improving robustness is not enough to handle the inherent class imbalance in accident datasets. To address this, we introduce a novel dynamic focal loss function designed to mitigate the bias toward non-accident events. Traditional focal loss functions help with class imbalance by reducing the weight of easily classified examples. Our dynamic variant goes a step further, adjusting the focus on difficult-to-classify samples on-the-fly, depending on the specific difficulty of each instance. This adjustment ensures that rare but crucial accident events receive the attention they need, while the model still learns to identify the more frequent non-accident scenarios. This dynamic, adaptive approach makes our model more effective at learning from both common and rare events, significantly improving the model's ability to anticipate accidents in real-world environments.

To simulate real-world challenges, we also incorporate Gaussian noise into our datasets to represent sensor failures and environmental disturbances. By injecting noise at varying levels of intensity, we simulate scenarios such as fog, heavy rain, or poor visibility, which are common in real-world driving conditions. This not only tests the robustness of our model but also trains it to handle incomplete or corrupted data. The ability to perform under these conditions is critical to ensure that accident anticipation models can be deployed effectively in actual autonomous driving systems.

We evaluate our framework on three widely-used datasets: DAD \cite{ref4}, CCD \cite{ref42}, and A3D \cite{ref1}. These datasets represent diverse, real-world traffic scenarios with varying levels of noise and sensor imperfections. Our extensive experiments show that our model outperforms existing state-of-the-art methods in terms of Average Precision (AP) and mean Time-to-Accident (mTTA), achieving higher accuracy and earlier detection times even under challenging conditions with missing or noisy data.

In summary, the contributions of this paper are as follows:\par
\begin{itemize}
    \item  We introduce Discrete Wavelet Transform (DWT) for multi-resolution feature extraction, which enhances the model’s resilience to sensor errors and environmental disturbances. This enables the model to handle both transient and steady traffic patterns, improving its adaptability to real-world conditions.
    
    \item  We propose a dynamic focal loss function that adjusts the focus on difficult-to-classify samples, addressing the class imbalance problem and improving the model's ability to predict rare accident events.
    
    \item We simulate sensor failures and environmental disturbances by introducing varying levels of Gaussian noise into the dataset. This augmentation ensures that our model is trained to handle imperfect data, demonstrating its robustness and ability to generalize to real-world conditions.
\end{itemize}

\section{RELATED WORK}
Traffic accident prediction has gained significant attention in recent years, largely driven by the rise of deep learning techniques. Traditional models often relied on statistical or rule-based approaches, but deep learning—especially with video-based traffic data—has proven more effective at capturing the complexity of traffic environments \cite{grant2018back}. Predicting accidents based on dashboard videos requires models that are not only accurate but also fast, as accidents can occur in just a few seconds. These models must be resilient to the inherent variability in traffic scenes, including unpredictable human behavior, varying weather conditions, and dynamic road layouts \cite{ref22}.

A core approach in accident prediction involves Convolutional Neural Networks (CNNs) \cite{ref4,ref9}, which are highly effective for feature extraction from traffic images and videos. However, CNNs struggle with capturing temporal dependencies, which are essential for understanding accident scenarios that evolve over time \cite{ref15,ref24,ref28}. To address this, sequential models such as Recurrent Neural Networks (RNNs), Long Short-Term Memory (LSTM) networks, and Gated Recurrent Units (GRUs) \cite{ref10,ref34,ref39} have been used to capture temporal sequences and the dynamic interactions between objects across frames, providing a better understanding of the evolving nature of traffic situations that lead to accidents \cite{ref40,ref43,ref44}.

Recent research has also explored the use of Graph Neural Networks (GNNs) \cite{ref18,ref26}, which model the complex relationships between multiple agents (e.g., vehicles, pedestrians) in traffic \cite{ref36,ref37,ref42}. GNNs are particularly useful for accident prediction because they capture spatial and temporal interactions, which are often key to determining the likelihood of an incident. Additionally, the transformer architecture \cite{ref11,ref38}, known for its ability to capture long-range dependencies, has been applied to improve understanding of spatial-temporal dynamics in traffic flow.

Another promising avenue is the use of generative models, such as Generative Adversarial Networks (GANs) and Variational Autoencoders (VAEs) \cite{ref2,ref41}, which generate realistic traffic scenarios for augmenting training data. This addresses the challenge of limited data for rare accidents, helping models learn from low-frequency events and generalize better to unseen situations.

Attention mechanisms have also proven to be effective in helping models focus on the most relevant features in traffic scenes. Spatial and temporal attention models \cite{ref17,ref18} enable the model to prioritize critical areas or time intervals that are more likely to cause accidents. These mechanisms have been shown to improve prediction accuracy by focusing on risky driving behaviors. Studies by Karim et al. \cite{ref17} and Song et al. \cite{ref32} demonstrated how attention models enhance both prediction accuracy and interpretability. Furthermore, Thakare et al. \cite{ref35} proposed a convolutional autoencoder for efficient feature extraction, while Liao et al. \cite{ref101} focused on capturing key scene elements to provide a broader understanding of traffic scenarios.

Despite advancements, real-world challenges persist in accident prediction, especially with noisy and incomplete data. Sensor failures, adverse weather, and road obstructions degrade model performance. Additionally, class imbalance, where accidents (positive samples) dominate over non-accident events, complicates predictions and leads to high false-positive rates. These issues remain underexplored in existing literature.

To address these, we propose the use of Discrete Wavelet Transform (DWT) for feature extraction, capturing both high- and low-frequency components to enhance robustness against noise and sensor degradation. We also apply a dynamic focal loss function to address class imbalance, ensuring that the model effectively learns from both frequent and rare events. Our approach improves accuracy and robustness by combining DWT for better feature extraction and dynamic focal loss for class imbalance, making it more suitable for real-world traffic prediction scenarios.

\section{METHODOLOGY}
\subsection{Problem Formulation}
\label{subsec1}

This work focuses on two primary objectives: estimating the likelihood of a traffic accident within a given time frame and detecting it as early as possible. We model the input as a sequence of $T$ frames from a dashboard video, denoted as $V = \{V_1, V_2, \dots, V_T\}$. Our task is to calculate the probability of an accident $\mathbf{P} = \{p_1, p_2, \dots, p_T\}$ for each frame.

If an accident occurs at time $t \in [1, T]$, the Time-to-Accident (TTA) with threshold $TTA_p$ is defined as $\Delta t = \tau - t_o$, where $\tau$ is the true time of the accident and $t_o$ is the first frame where the probability $p_t$ exceeds a predefined threshold $p$.

A video is classified as containing an accident (positive) if there exists a frame $t \geq t_o$ such that $p_t \geq p_o$, and as not containing an accident (negative) when $\tau = 0$. The primary goal of our model is to improve the accuracy of accident detection and maximize the time-to-accident, enabling the earliest possible prediction of a potential collision.

\subsection{Framework Overview}
\label{subsec2}

\begin{figure}
\centering
\includegraphics[width=1\textwidth]{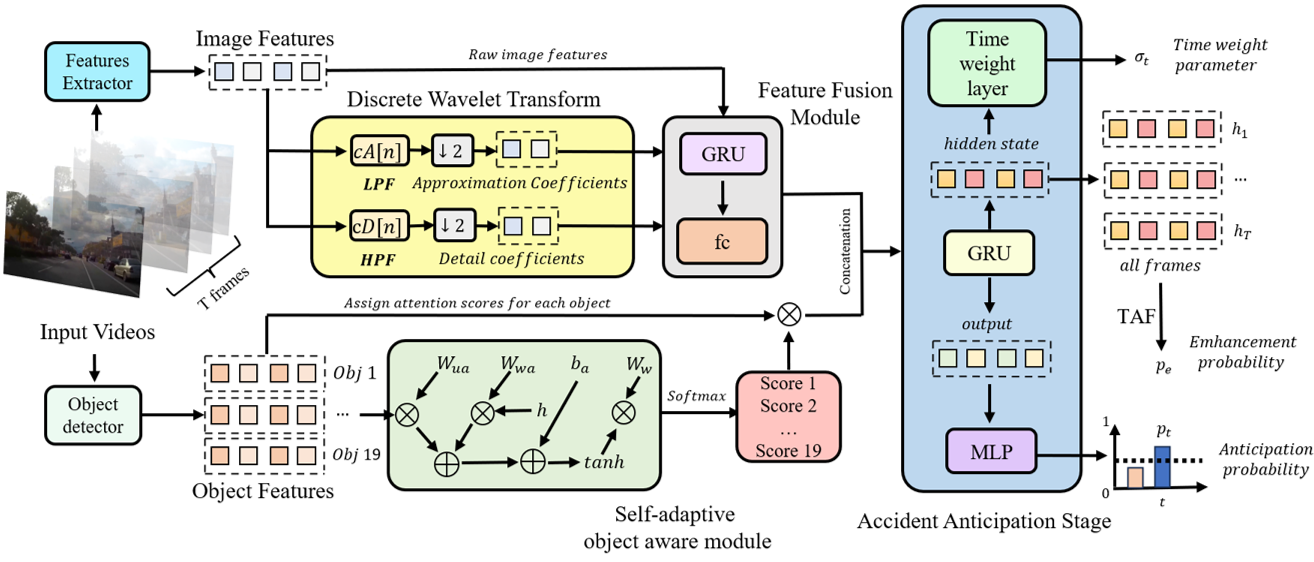}
\caption{Overview of the proposed framework.
The input video frames are processed by an object detector and feature extractor to obtain object-level and image-level features. These features are refined through a Self-Adaptive Object-Aware Module and Discrete Wavelet Transform (DWT). The refined features are fused and passed through a GRU and Temporal Attention Fusion to compute the anticipation probability ($p_t$) and enhancement probability ($p_e$), with a time weight layer adjusting temporal influence on predictions. The framework integrates spatial, temporal, and hierarchical features for enhanced prediction accuracy.}
\label{tab:Fig. 1}
\end{figure}

The overall pipeline of ROAR is illustrated in Figure \ref{tab:Fig. 1}, consisting of five key components: object detector, feature extractor, self-adaptive object-aware module, discrete wavelet transformation module, and temporal attention fusion. The pipeline begins with the object detector and feature extractor, which generate object and image vectors from the raw input video. The object vectors are then processed by the self-adaptive object-aware module, which updates the spatial-temporal representation and produces the object-aware vectors \( \mathbf{\bar{F}}_{obj} \). Meanwhile, the image vectors \( \mathbf{F}_{img} \) are passed through the discrete wavelet transformation module to decompose the features into approximation and detail components, yielding the approximation coefficients \( c_A \) and detail coefficients \( c_D \). These components are then fused in the feature fusion module to produce the final image vectors \( \mathbf{\bar{F}}_{img} \). 

The fused features are passed to the GRU module, which iteratively combines and refines the outputs from the feature extractor and the other modules. This allows the model to identify and predict potential incidents leading to accidents, generating the probability \( p_t \) for each frame. To refine the predictions, the time weight layer and temporal attention fusion are applied to account for hidden temporal dependencies.

\paragraph{Object Detector}
Given a sequence of \( T \) frames from a dashboard video, we use Cascade R-CNN \cite{ref3} to detect dynamic objects such as cars, motorcycles, and pedestrians. The top \( T \) objects are selected based on their recognition scores and embedded into object vectors \( \mathbf{F}_{obj} \) using VGG-16 \cite{ref31}.

\paragraph{Feature Extractor}
The feature extractor captures semantic features from the entire video sequence. It employs both VGG-16 and a multilayer perceptron (MLP) to generate the image vectors \( \mathbf{F}_{img} \), which represent the underlying features of the video content.

\paragraph{Self-Adaptive Object-Aware Module}
The self-adaptive object-aware module enhances spatial representations by applying attention weights to different objects based on their interactions with hidden states and temporal context. The attention mechanism is computed as follows:

\begin{equation}
\mathbf{e}_t = \text{tanh}(\mathbf{h}_{t-1} \mathbf{W}_{wa} + \mathbf{F}_{obj} \mathbf{W}_{ua} + \mathbf{b}_a)
\end{equation}
where \( \mathbf{W}_{ua} \) and \( \mathbf{W}_{wa}  \) are learnable weight matrices, \( \mathbf{b}_a  \) is the bias term, and \( \mathbf{h}_{t-1} \) is the hidden state at time step \( t-1 \). To compute the attention weights \( \alpha \), we apply the softmax function to the sum of the attention energies over the objects. First, a broadcasted version of the weights \( \mathbf{W}_w \) is applied: 

\begin{equation}
\mathbf{e}_t' = \mathbf{e}_t \mathbf{W}_w
\end{equation}
The attention weights \( \alpha \) are then computed by applying softmax over the summed energies for all objects:

\begin{equation}
\alpha_t = \text{softmax} \left(  \mathbf{e}_t' \right)
\end{equation}
where \( \alpha_t \) represents the normalized attention weights for each object at time step \( t \). The attention weights \( \alpha_t \) are applied to the object embeddings through element-wise multiplication. The updated object embeddings are computed as:

\begin{equation}
\mathbf{\bar{F}}_{obj} = \mathbf{\alpha}_t \cdot \mathbf{F}_{obj}
\end{equation}
Finally, the updated object embeddings \(\mathbf{\bar{F}}_{obj}\) are returned as the output of the Self-adaptive object-aware module. This method allows the model to selectively focus on relevant objects and incorporate both spatial and temporal dependencies, making the attention mechanism adaptive to each frame in the sequence. The learned weights \( \mathbf{W}_{ua}, \mathbf{W}_{wa}, \mathbf{W}_w \) and the bias term \( \mathbf{b}_a \) ensure that the attention mechanism adjusts dynamically based on the input data, effectively modeling spatial interactions in the scene.

\paragraph{Discrete Wavelet Transformation Module}
The Discrete Wavelet Transform (DWT) is applied to the image feature vectors to capture both low-frequency (approximation) and high-frequency (detail) components. This decomposition helps the model capture both smooth, coarse-grained features as well as fine-grained, detailed variations in the data. We use 1-D DWT for feature vectors, as some of the original video data in datasets has been deleted by publisher.

\begin{figure}
\centering
\includegraphics[width=0.8\textwidth]{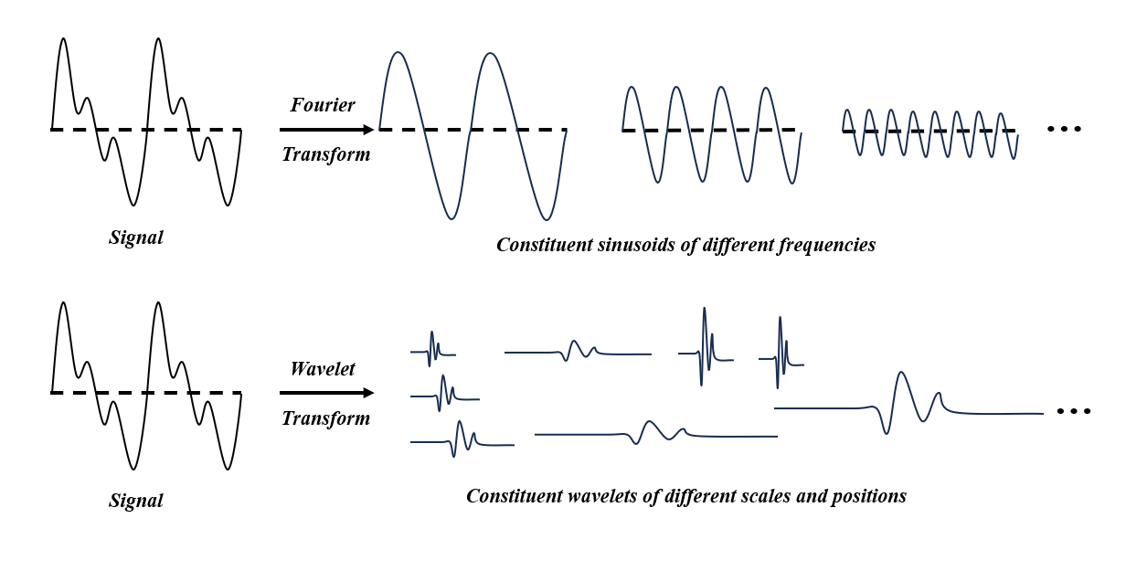}
\caption{Comparison between Fourier Transform and Wavelet Transform. The upper panel illustrates how the Fourier Transform decomposes a signal into constituent sinusoids of different frequencies, which represent the signal in the frequency domain. The lower panel demonstrates how the Wavelet Transform decomposes a signal into wavelets of different scales and positions, offering a time-frequency representation that preserves both time and frequency information. This comparison highlights the advantages of Wavelet Transform in capturing localized features of the signal, particularly for non-stationary or time-varying data.}
\label{tab:Fig. 2}
\end{figure}

Mathematically, the wavelet transform decomposes a signal \( f(x) \) into a sum of scaled and shifted versions of a mother wavelet function \( \psi(x) \) and a scaling function \( \phi(x) \). This decomposition can be expressed as:

\begin{equation}
f(x) = \sum_{k} cA_k \phi(x - k) + \sum_{k} cD_k \psi(x - k)
\end{equation}
where \( \phi(x) \) represents the scaling function, which captures the low-frequency (approximation) part of the signal, while \( \psi(x) \) is the wavelet function that captures the high-frequency (detail) part of the signal. The coefficients \( cA_k \) correspond to the approximation coefficients, which capture the smooth (low-frequency) components of the signal, and \( cD_k \) are the detail coefficients, which capture the high-frequency (fine-grained) components. The parameter \( k \) serves as the translation factor that shifts both the scaling and wavelet functions.

Building on the decomposition of the signal \( f(x) \), we now focus on the two main components that the transform captures: the approximation coefficients and the detail coefficients. These components allow the signal to be represented at multiple scales, with each capturing different aspects of the frequency content.

\textbf{Approximation Coefficients} \( cA_k \): These coefficients, related to the scaling function \( \phi(x) \), represent the low-frequency, smooth components of the signal. They capture the general trends and coarse structures over time. The approximation coefficients are computed as:

\begin{equation}
cA_k = \langle f(x), \phi(x - k) \rangle
\end{equation}
where \( \langle f(x), \phi(x - k) \rangle \) denotes the inner product between the signal \( f(x) \) and the shifted scaling function \( \phi(x - k) \), which effectively extracts the low-frequency content.

\textbf{Detail Coefficients} \( cD_k \): These coefficients, related to the wavelet function \( \psi(x) \), capture the high-frequency components of the signal. They describe the fine-grained variations and changes over short time intervals, highlighting the detailed structure of the signal. The detail coefficients are calculated as:

\begin{equation}
cD_k = \langle f(x), \psi(x - k) \rangle
\end{equation}
where the inner product between \( f(x) \) and the shifted wavelet function \( \psi(x - k) \) extracts the high-frequency components, capturing the rapid changes in the signal.

Together, the approximation and detail coefficients provide a comprehensive multi-resolution representation of the signal, allowing the wavelet transform to capture both smooth, large-scale patterns and fine, high-frequency details. This multi-scale decomposition is particularly useful for analyzing signals with non-stationary behavior, as it enables a richer representation of both coarse and detailed features.

In this work, the Daubechies wavelet with 1 vanishing moment (denoted as db1) is used for the transformation. The Daubechies wavelets are a family of orthogonal wavelets that have compact support and are widely used in signal processing for their good time-frequency localization properties. For the db1 wavelet (also known as the Haar wavelet), the scaling and wavelet functions are simple step functions, making it computationally efficient for applications. The db1 wavelet can be expressed as:

\begin{equation}
\phi(x) = \begin{cases} 
1 & \text{if } 0 \leq x < 1, \\
0 & \text{otherwise}.
\end{cases}
\end{equation}

\begin{equation}
\psi(x) = \begin{cases} 
1 & \text{if } 0 \leq x < 0.5, \\
-1 & \text{if } 0.5 \leq x < 1, \\
0 & \text{otherwise}.
\end{cases}
\end{equation}

The simplicity of the Haar wavelet functions makes it computationally efficient, allowing for rapid transformations. This simplicity is particularly beneficial for real-time applications and environments with limited computational resources.

\begin{figure} [htbp]
\centering
\includegraphics[width=0.7\textwidth]{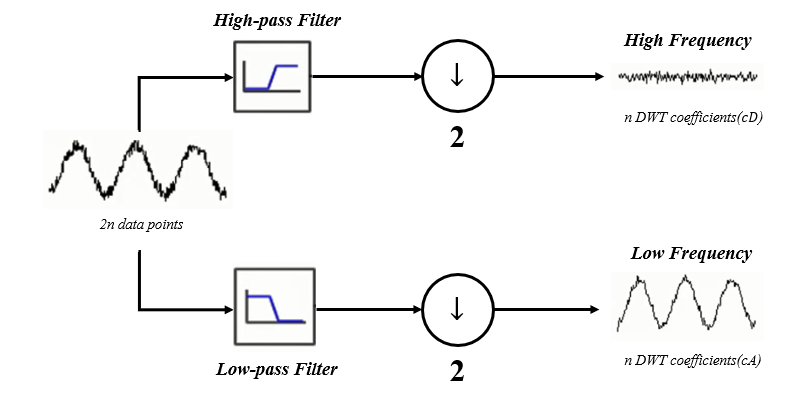}
\caption{One-level Discrete Wavelet Transform (DWT) decomposition applied to discrete data points. The process involves filtering the data through a high-pass filter to extract high-frequency components (denoted as $cD$) and through a low-pass filter to extract low-frequency components (denoted as $cA$). The original data is divided into two halves, with the high-pass filter focusing on capturing the rapid variations (high-frequency) and the low-pass filter capturing the smooth trends (low-frequency).}
\label{tab:Fig. 3}
\end{figure}

For continuous wavelet functions, the above forms are applicable. However, when working with discrete input features, the scaling function $\phi(x)$ and the wavelet function $\psi(x)$ must also be discretized. For the Haar wavelet, both the discretized wavelet and scaling functions can be directly derived from their continuous counterparts.

The discretized scaling function for Haar can be expressed as:

\begin{equation}
\phi[n] =
\begin{cases}
1 & \text{if } n = 0,1, \\
0 & \text{otherwise}.
\end{cases}
\end{equation}

Similarly, the discretized wavelet function is defined as:

\begin{equation}
\psi[n] =
\begin{cases}
1 & \text{if } n = 0, \\
-1 & \text{if } n = 1, \\
0 & \text{otherwise}.
\end{cases}
\end{equation}

Once discretized, the DWT is computed via convolution. For a given discrete signal $f[n]$, the DWT can be expressed as:

\begin{equation}
cA[k] = \sum_n f[n] \phi[n - 2k]
\end{equation}
\begin{equation}
cD[k] = \sum_n f[n] \psi[n - 2k]
\end{equation}
where $cA[k]$ represents the approximation coefficients and $cD[k]$ represents the detail coefficients. The factor $2k$ accounts for the shift of the scaling and wavelet functions, as the Haar wavelet is binary, with both the scaling and wavelet functions having a length of 2.

In our model, the DWT is applied to each feature vector \( \mathbf{F}_{img} \) at time step \( t \). After downsampling the feature vectors, the DWT captures both the low-frequency and high-frequency components, providing a detailed representation of the signal. The transformation produces approximation coefficients \( cA \) and detail coefficients \( cD \), which are combined into a new vector \( \mathbf{C}_{combined} \), stored in the output tensor:

\begin{equation}
\mathbf{C}_{combined} = \text{concat}(cA, cD)
\end{equation}

Once the wavelet-transformed features are obtained, they are fused with the original image features. A GRU (Gated Recurrent Unit) network is used to combine the original image tensor \( \mathbf{F}_{img} \) and the combined wavelet coefficients \( \mathbf{C}_{combined} \). The GRU, followed by a fully connected layer (fc), processes these tensors and generates the fused feature vector \( \mathbf{\bar{F}}_{img} \), which integrates both spatial and temporal aspects of the input data:

\begin{equation}
\mathbf{\bar{F}}_{img}  = \text{fc}(\text{GRU}(\mathbf{F}_{img}, \mathbf{C}_{combined}))
\end{equation}

The next step involves fusing the image features \( \mathbf{\bar{F}}_{img} \) with the object-aware features \( \mathbf{\bar{F}}_{obj} \), which are then passed through a second GRU network. This GRU processes the concatenated features and generates the hidden state \( \mathbf{h_t} \) as well as the final output \( \mathbf{X_t} \) for the \( t \)-th frame of the video:

\begin{equation}
\mathbf{X_t}, \mathbf{h_t} = \text{GRU}(\text{concat}(\mathbf{\bar{F}}_{img}, \mathbf{\bar{F}}_{obj}))
\end{equation}

The accident probability score \( p_t \) for the \( t \)-th frame is then predicted using a Multi-Layer Perceptron (MLP):

\begin{equation}
{p_t} = \text{MLP}(\mathbf{X_t})
\end{equation}

At the same time, the time weight parameter \( \sigma_t \) is generated by applying a time weight layer, which utilizes a fully connected (fc) layer:$\sigma_t = \text{time\_weight\_layer}(\mathbf{h_t})$. To better capture the temporal dependencies over the sequence, the temporal attention fusion module is introduced. This module aggregates features from across the timeline by first performing both max pooling and average pooling over \( \mathbf{h_t} \). The pooled features are then concatenated to form the aggregated spatial features:

\begin{equation}
F_{agg} = \text{concat}(\text{avgpool}, \text{maxpool})
\end{equation}
Next, the attention energy is computed by calculating the pairwise similarity between the aggregated spatial features:

\begin{equation}
\mathbf{E} = F_{agg}^T \cdot F_{agg}
\end{equation}
The attention weights \( \alpha \) are derived by applying the softmax function, i.e., $\alpha = \text{softmax}(\mathbf{E})$. 
These attention weights are then used to weight the aggregated spatial features:

\begin{equation}
\mathbf{W} = F_{agg} \cdot \alpha
\end{equation}
Finally, the weighted features are projected to the final feature dimension using a learned weight vector \( \mathbf{w} \):

\begin{equation}
\mathbf{A} = \mathbf{W}^T \cdot \mathbf{w}
\end{equation}
The output of the temporal attention fusion module, \( \mathbf{A} \), represents the final aggregated feature that combines temporal and spatial information. A final fully connected layer (fc) estimates the auxiliary loss \( p_a \):

\begin{equation}
{p_a} = \text{fc}(\mathbf{A})
\end{equation}

\subsection{Training Loss}
\label{subsec1}
In this section, we define the loss function used to train the model, which combines exponential loss and dynamic focal loss to effectively address the imbalanced binary classification task. The total anticipation loss function consists of two parts: positive and negative losses, with each part weighted according to both the network's output and the temporal dynamics of the task. We first introduce the dynamic focal loss component.

\paragraph{Dynamic Focal Loss}
The dynamic focal loss is specifically designed to handle class imbalance by focusing more on difficult-to-classify examples. This loss is computed as:

\begin{equation}
\mathcal{L}_{\text{focal}} = \alpha (1 - p_c)^\gamma \cdot \mathcal{L}_{\text{ce}}
\end{equation}
where \( \alpha \) is the weight for positive examples, dynamically adjusted to handle imbalanced data, and \( p_c = \exp(-\mathcal{L}_{\text{ce}}) \) is the predicted probability for the correct class. The parameter \( \gamma \) is the focusing parameter, typically set to 2, which increases the loss for misclassified examples and directs the model’s attention to harder-to-classify samples. \( \mathcal{L}_{\text{ce}} \) represents the standard cross-entropy loss.

\paragraph{Loss for Positive Samples}
For positive samples, we apply a penalty based on the temporal difference between the predicted time of accident  and the actual time of accident. This penalty is computed as:

\begin{equation}
\text{penalty} = -\max\left(0, \frac{\text{toa} - \text{time} - 1}{\text{fps}}\right)
\end{equation}
This temporal penalty adjusts the loss depending on the predicted and actual timing of the accident. The time weight is computed using the hidden state \( h \) from the GRU and a time weight layer:

\begin{equation}
\text{time\_weight} = 1 + \sigma\left(\text{time\_weight\_layer}(h)\right)
\end{equation}
where \( \sigma(\cdot) \) is the sigmoid function. The time-weighted exponential loss for positive examples is then given by:

\begin{equation}
\mathcal{L}_{\text{pos}} = -\left(\text{time\_weight} \cdot \exp(\text{penalty}) \cdot \mathcal{L}_{\text{focal}}\right)
\end{equation}
where \( \mathcal{L}_{\text{focal}} \) is the dynamic focal loss for positive samples, combined with the time-weighted factor and temporal penalty.

\paragraph{Loss for Negative Samples}
For negative samples, we apply the standard cross-entropy loss, multiplied by a constant \( c \) to adjust its contribution:

\begin{equation}
\mathcal{L}_{\text{neg}} = c \mathcal{L}_{\text{ce}}
\end{equation}

\paragraph{Total Loss}
The total anticipation loss is a weighted sum of the positive and negative losses, where the weights depend on the binary target values (either 0 or 1):
\begin{equation}
\mathcal{L}_{an} = \mathbb{E}\left[\sum_{i=1}^{T} \left( \alpha_i \cdot \mathcal{L}_{\text{pos}}(i) + (1 - \alpha_i) \cdot \mathcal{L}_{\text{neg}}(i) \right)\right]
\end{equation}
In addition to the anticipation loss, we also compute an auxiliary loss. This is the cross-entropy loss between the predicted auxiliary output \( p_a \) and the true labels \( y \), averaged over the batch:

\begin{equation} \mathcal{L}_{au} = \frac{1}{T} \sum_{i=1}^{T} \mathcal{L}_{\text{ce}}(p_e, y_i)
\end{equation}

Finally, the total loss is a linear combination of the anticipation loss and the auxiliary loss:

\begin{equation}
\mathcal{L}_{total}=  \mathcal{L}_{an} + \beta \mathcal{L}_{au}
\end{equation}
where \( \beta \) is a hyperparameter that determines the relative importance of the auxiliary loss \( \mathcal{L}_{au} \) in the overall training process. By adjusting \( \beta \), we can control the balance between the anticipation loss and the auxiliary loss to better suit the specific task and improve the model’s performance.

\section{EXPERIMENT}
\label{4}
\subsection{Experiment Setup}
In our experiments, we evaluate the performance of our model using three distinct datasets: the Dashcam Accident Dataset (DAD) \cite{ref4}, the Car Crash Dataset (CCD) \cite{ref42}, and the AnAn Accident Detection (A3D) \cite{ref1} dataset. These datasets collectively provide a diverse range of real-world traffic accident scenarios, offering valuable insights into accident detection across various environments and conditions.

\begin{itemize}
    \item \textbf{Dashcam Accident Dataset (DAD):} The DAD comprises 620 positive clips (accident-containing) and 1,130 negative clips (accident-free), each containing 100 frames over 5 seconds. Captured at 720p resolution across six major Taiwanese cities, the dataset includes multiple accident types, such as car-motorcycle, car-to-car, and motorcycle-to-motorcycle incidents. This diversity enables the training and evaluation of accident anticipation models under various conditions, including edge cases.

    \item \textbf{Car Crash Dataset (CCD):} The CCD comprises 1,500 positive clips and 3,000 negative clips, each containing 50 frames over 5 seconds. These clips are divided into 3,600 training clips and 900 testing clips. The dataset offers detailed annotations, including ego-vehicle involvement, accident participants, and causal mechanisms. This comprehensive annotation facilitates a nuanced analysis of accident scenarios, supporting the development of models that can anticipate accidents under varied conditions.

    \item \textbf{AnAn Accident Detection (A3D) Dataset:} The A3D dataset comprises 1,087 positive clips and 114 negative clips, each 5 seconds long with 100 frames. The clips are divided into 961 training clips and 240 testing clips. Documenting abnormal road events across East Asian urban environments, A3D maintains temporal and structural configurations identical to DAD, making it suitable for traffic accident anticipation research.
\end{itemize}

To further challenge the model and assess its robustness, we enhance the A3D dataset by introducing Gaussian noise with varying variances. This step is designed to simulate the effects of environmental noise and sensor imperfections that commonly occur in real-world settings, particularly in challenging conditions such as fog, heavy rain, or low-light environments. By incorporating noise of different intensities, we create more complex and unpredictable scenarios, encouraging the model to learn more resilient features. This augmentation process improves the model’s ability to generalize under noisy conditions, ultimately making it more adaptable to real-world variations and enhancing the overall reliability of accident detection systems in practical applications.

\subsection{Evaluation Metrics}

In this study, we evaluate the model's performance using two key metrics: accuracy and timeliness. These metrics provide a comprehensive view of the model's effectiveness in both correctly identifying accidents and predicting them promptly.

\textbf{Accuracy.} The model’s effectiveness in detecting accidents is assessed through recall (\(R\)), which quantifies the proportion of correctly identified accidents (true positives, TP) relative to all actual accidents (TP + false negatives, FN). Precision (\(P\)) measures the model's reliability, defined as the ratio of true positives (TP) to the total number of predicted positive instances (TP + false positives, FP). However, since recall and precision often vary with different threshold settings, we use Average Precision (AP) as a more comprehensive metric. AP is computed as the area under the precision-recall curve:

\begin{equation}
AP = \int P(R) \, dR
\end{equation}
This metric provides an overall evaluation of the model's ability to consistently predict accidents with high accuracy across various thresholds. A higher AP value indicates superior model performance and robustness across different operating points.

\textbf{Timeliness.} Timeliness in accident detection is assessed using the Time-to-Accident (TTA), which measures the time difference between the model's prediction of an impending accident (once the predicted risk exceeds a specified threshold) and the actual occurrence of the accident. A longer TTA implies that the model can detect potential accidents earlier, thus providing more time for preventive actions. The Mean Time-to-Accident (mTTA) is the average TTA across multiple thresholds (\(a\)), reflecting the model's overall lead time in predicting accidents:

\begin{equation}
\text{mTTA} = \int_0^1 \text{TTA}_ a\, da
\end{equation}

Additionally, we calculate TTA@R80, which measures the TTA when the model achieves a recall rate of 80\%. This metric emphasizes the model’s effectiveness in delivering early warnings, particularly under conditions where the recall is high.

\subsection{Implementation Details}
The proposed model is implemented using PyTorch 2.0 and trained on an NVIDIA GeForce RTX 3050 GPU. The training process spans 30 epochs, with a consistent batch size of 10. We utilize the Adam optimizer for gradient descent and employ the ReduceLROnPlateau strategy for learning rate adjustment. The initial learning rate is set to $1 \times 10^{-3}$, which is uniformly applied across all datasets.

For object detection, the model is configured to detect up to 19 candidate objects. The VGG-16 network is used for feature extraction, with an embedding dimension of 4096. Additionally, the GRU (Gated Recurrent Unit) network is employed, with a hidden state dimension fixed at 128.

\begin{table}[ht]
    \centering
    \small
    \begin{tabular}{lcccccc}
        \toprule
        \multirow{2}{*}{\textbf{Model}} & \multicolumn{2}{c}{\textbf{DAD}} & \multicolumn{2}{c}{\textbf{CCD}} & \multicolumn{2}{c}{\textbf{A3D}} \\
        \cmidrule(lr){2-3} \cmidrule(lr){4-5} \cmidrule(lr){6-7}
        & \textbf{AP (\%)}$\uparrow$ & \textbf{mTTA (s)}$\uparrow$ & \textbf{AP (\%)}$\uparrow$ & \textbf{mTTA (s)}$\uparrow$ & \textbf{AP (\%)}$\uparrow$ & \textbf{mTTA (s)}$\uparrow$ \\
        \midrule
        DSA\cite{ref4} &48.1 & 1.34 & 98.7 & 3.08 & 92.3 & 2.95 \\
        ACRA\cite{ref45} & 51.4 & 3.01 & 98.9 & 3.32 & - & - \\
        AdaLEA\cite{ref33} & 52.3 & 3.44 & 99.2 & 3.45 & 92.9 & 3.16 \\
        UString\cite{ref17} & 53.7 & 3.53 & \underline{99.5} & \underline{3.74} & 93.2 & 3.24 \\
        DSTA\cite{ref1} & 52.9 & 3.21 & 99.1 & 3.54 & 93.5 & 2.87 \\
        GSC\cite{ref37} & 58.2 & 2.76 & 99.3 & 3.58 & 94.9 & 2.62 \\
        AccNet \cite{ref100} & \underline{60.8} & \underline{3.58} & \underline{99.5} & \textbf{3.78} & \underline{95.1} & \underline{3.26} \\
        \cellcolor{purple!15}\textbf{ROAR} & \cellcolor{purple!15}\textbf{76.0} & \cellcolor{purple!15}\textbf{3.92} & \cellcolor{purple!15}\textbf{99.8} & \cellcolor{purple!15}3.24 & \cellcolor{purple!15}\textbf{95.3} & \cellcolor{purple!15}\textbf{3.73} \\
        \bottomrule
    \end{tabular}
    \caption{Comparison of model performance in balancing mTTA and AP across three datasets. The best and second-best performance values in each category are highlighted in bold and underlined, respectively. Instances where data is unavailable are marked with a dash (“-”). The arrows indicate that higher values for AP (\%) and mTTA (s) are better.}
\label{tab:table1}
\end{table}

\subsection{Evaluation Results}
\noindent\textbf{Compare with SOTA Baselines on Raw Datasets.} As shown in Table \ref{tab:table1}, our model demonstrates clear advantages over SOTA baselines across three dataset in both accuracy (AP) and timeliness (mTTA), delivering consistent improvements across all datasets.

On the DAD and A3D datasets, our model not only achieves high accuracy but also provides timely accident predictions. Specifically, on the DAD dataset, our model reaches an AP of 75.9\%, surpassing the next best baseline, AccNet, by 15.1\%. The model also achieves an mTTA of 3.91 seconds, highlighting its ability to offer both accurate and timely accident predictions.

These results emphasize the model’s robustness in real-world traffic conditions, where both predictive accuracy and early detection are crucial for accident prevention. The timeliness of the model’s alerts is particularly important, as demonstrated by its lead time, which enables earlier intervention.

Further analysis on the DAD dataset, as shown in Table \ref{tab:table2}, reveals that our model not only achieves the highest AP but also provides the earliest possible warning, with an average lead time of 3.72 seconds before an accident occurs. This early lead time is crucial for real-time accident avoidance systems, providing drivers with more time to react to potential hazards.

In conclusion, our model sets a new benchmark in accident prediction by balancing high accuracy with timely warnings. These advancements make it a significant improvement over existing methods, underscoring its potential for deployment in real-world traffic safety systems.

\begin{table}[ht]
    \centering
    \small
    \begin{tabular}{lccc}
        \toprule
        \textbf{Model} & \textbf{AP (\%)}$\uparrow$ & \textbf{mTTA (s)}$\uparrow$ & \textbf{@R80 (s)}$\uparrow$ \\
        \midrule
        DSA \cite{ref4} & 63.50 & 1.67 & 1.85 \\
        DSTA \cite{ref1} & 66.70 & 1.52 & 2.39 \\
        UString \cite{ref17} & 68.40 & 1.63 & 2.18 \\
        GSC \cite{ref37} & 68.90 & 1.33 & 2.14 \\
        AccNet \cite{ref100} & 70.10 & 1.73 & 2.23 \\\cellcolor{purple!15}\textbf{ROAR} & \cellcolor{purple!15}\textbf{79.53} & \cellcolor{purple!15}\textbf{3.72} & \cellcolor{purple!15}\textbf{3.62} \\
        \bottomrule
    \end{tabular}
    \caption{Comparison of models based on the best AP performance on the DAD dataset. The column TTA@R80 shows the Mean Time-to-Accident (mTTA) at a recall level of 80\%.}
    \label{tab:table2}
\end{table}

\begin{table}[ht]
    \centering
        \small
    \begin{tabular}{ccc}
        \toprule
        \textbf{Variance ($\sigma^2$)} & \textbf{AP (\%)}$\uparrow$ & \textbf{mTTA (s)}$\uparrow$  \\
        \midrule
        \cellcolor{purple!15}\textbf{ROAR (Original)} & \cellcolor{purple!15}\textbf{95.3} & \cellcolor{purple!15}\textbf{3.73} \\
        + 0.1 & 94.6 & 3.27 \\
        + 0.2 & 94.6 & 3.16 \\
        + 0.5 & 94.9 & 3.32 \\
        + 1.5 & 93.9 & 3.99 \\
        + 5.0 & 93.5 & 3.33 \\
        + 20.0 & 92.3 & 3.32 \\
        \bottomrule
    \end{tabular}
    \caption{Comparison of different levels of Gaussian noise variance and their impact on balancing mTTA and AP on the A3D dataset. "Original" refers to the baseline performance without any added noise.}
    \label{tab:table3}
\end{table}

\noindent\textbf{Performance on Datasets with Gaussian Noise.} 
Table \ref{tab:table3} illustrates the robustness of our model in handling noise. In this experiment, Gaussian noise with a mean of 0 and varying standard deviations is introduced to simulate real-world conditions. Despite the presence of noise in the A3D dataset, our model consistently outperforms nearly all other models. Notably, when Gaussian noise with a variance of 0.5 is added, our model maintains a strong Average Precision (AP) of 94.9\%. This demonstrates the model’s ability to preserve its predictive accuracy even under noisy conditions. 

As expected, the model's performance does degrade with increasing noise, but it remains resilient. Even with higher noise levels (e.g., variance = 5), our model continues to perform competitively, maintaining high predictive accuracy compared to models trained on clean, noise-free data. These results underscore the model's robustness, highlighting its suitability for real-world driving environments, where sensor noise and data imperfections are inevitable.

\subsection{Ablation Studies}
\noindent\textbf{Ablation Study for Core Components.} Table \ref{tab:table4} presents the results of our ablation study, which systematically evaluates the contribution of five core components in the ROAR model: Discrete Wavelet Transform (DWT), self-adaptive object aware module, temporal attention fusion, time weight layer, and dynamic focal loss. Evaluating the model on the A3D dataset, we observe that the removal of any of these components leads to a significant degradation in performance, as evidenced by notable decreases in both AP and mTTA metrics when compared to the full model.

Among the most influential components, the combination of DWT and dynamic focal loss stands out for its substantial impact. The DWT enhances feature extraction by capturing both low and high-frequency components, allowing the model to effectively handle noisy and incomplete data. The inclusion of dynamic focal loss helps the model address class imbalance by emphasizing the harder-to-classify accident instances, improving both the detection accuracy and prediction timeliness. Specifically, these two components contribute a 2.9\% increase in AP and a 0.64-second improvement in mTTA. These results show that the DWT's ability to capture multi-scale features and dynamic focal loss’s ability to focus on rare accident events play pivotal roles in improving both prediction accuracy and the model’s ability to anticipate accidents earlier, which is crucial for real-time applications in autonomous driving.

\begin{table}[ht]
    \centering
        \small
    \begin{tabular}{l|c|c}
        \hline
        \textbf{Experiment} & \textbf{AP (\%)}$\uparrow$ & \textbf{mTTA (s)}$\uparrow$ \\
        \hline
        \cellcolor{purple!15}\textbf{ROAR (Full Model)} & \cellcolor{purple!15}\textbf{95.3} & \cellcolor{purple!15}\textbf{3.73} \\
        w/o Discrete Wavelet Transform & 92.6 & 3.09 \\
        w/o Discrete Wavelet Transform & 92.6 & 3.09 \\
        w/o Self-adaptive Object Aware Module & 91.9 & 3.06 \\
        w/o Temporal Attention Fusion & 93.8 & 3.49 \\
        w/o Time Weight Layer & 95.0 & 3.82 \\
        w/o Dynamic Focal Loss & 90.1 & 3.51 \\
        \hline
    \end{tabular}
    \caption{Ablation studies of different modules on A3D dataset. "w/o" denotes removal of a module.}
    \label{tab:table4}
\end{table}

When the self-adaptive object-aware module is removed, the model’s AP decreases to 91.9\%, and mTTA increases to 3.06 seconds, demonstrating that this module plays a key role in identifying and prioritizing high-risk vehicles. By focusing on these vehicles and analyzing their spatial-temporal relationships, the model can predict accidents more reliably. Similarly, when temporal attention fusion is excluded, the model’s mTTA increases to 3.49 seconds, and AP drops to 93.8\%. Temporal attention fusion is vital for understanding the temporal dependencies between traffic agents, allowing the model to recognize patterns that evolve over time and improve the model’s timeliness in anticipating accidents.

Interestingly, the time weight layer leads to a slight improvement in mTTA, from 3.73 seconds to 3.82 seconds, when removed. While this may seem counterintuitive, it suggests that the model still maintains good performance even without explicitly weighing the temporal sequence. However, the absence of the time weight layer reduces the model’s ability to fine-tune its focus on the most critical temporal instances, which could help reduce the prediction lead time.

These results underscore the importance of each component in improving the overall performance of the ROAR model. Each element contributes to the model’s ability to handle real-world challenges, such as noisy data, class imbalance, and the need to prioritize high-risk traffic agents. The removal of any component results in a noticeable degradation of both accuracy and timeliness, emphasizing the need for a holistic approach.

\begin{table}[htbp]
    \centering
        \small
    \begin{tabular}{ccccc}
        \toprule
        \textbf{Approximation} & \textbf{Details} & \textbf{Features fusion} & \textbf{AP(\%)}$\uparrow$ & \textbf{mTTA(s)}$\uparrow$ \\
        \midrule
        \cellcolor{purple!10}$\surd$ & \cellcolor{purple!10}$\surd$ & \cellcolor{purple!10}$\surd$ & \cellcolor{purple!10}\textbf{95.3} & \cellcolor{purple!10}\textbf{3.73} \\
        $\surd$ & $\surd$ & $\times$ & 90.5 & 3.28 \\
        $\surd$ & $\times$ & $\surd$ & 93.3 & 3.07 \\
        $\surd$ & $\times$ & $\times$ & 91.6 & 3.43 \\
        $\times$ & $\surd$ & $\surd$ & \underline{94.6} & \underline{3.51} \\
        $\times$ & $\surd$ & $\times$ & 91.1 & 3.11 \\
        \bottomrule
    \end{tabular}
    \caption{Ablation studies of approximation and details of Discrete Wavelet Transform and features fusion module on A3D dataset. $\surd$ represents inclusion, and $\times$ represents exclusion of the respective module.}
    \label{tab:table5}
\end{table}

\noindent\textbf{Impact of DWT and Feature Fusion.} In addition to the core component analysis, we further investigate the effect of the Discrete Wavelet Transform (DWT) by evaluating how the approximation coefficients (\(cA\)) and detail coefficients (\(cD\)) contribute individually and in combination to the model’s performance. As shown in Table \ref{tab:table5}, different configurations were tested, such as using only the approximation coefficients (\(cA\)), only the detail coefficients (\(cD\)), or both, with and without fusion with the raw features.

The results indicate that using both \(cA\) and \(cD\) in conjunction with the raw features results in the highest performance (95.3\% AP and 3.73 seconds mTTA), demonstrating the power of combining multi-resolution information. The approximation coefficients capture the broad, low-frequency aspects of the traffic scene, while the detail coefficients highlight sudden, high-frequency changes that are indicative of potential accidents. The fusion of these two components provides a comprehensive view of the traffic environment, enhancing the model’s ability to predict accidents in both stable and dynamic conditions.

When only one set of coefficients is used, the model performance decreases. Using only \(cA\) yields an AP of 91.6\% and an mTTA of 3.43 seconds, while using only \(cD\) results in an AP of 94.6\% but a slightly slower mTTA of 3.51 seconds. These findings suggest that while each set of coefficients contributes valuable information, the combination of both provides a richer feature set that better captures the full spectrum of traffic dynamics. Furthermore, the fusion with raw features further strengthens the model's predictive capabilities, underscoring the importance of leveraging both the fine and coarse details from the data.

In conclusion, the ablation study highlights the critical role that both the approximation and detail coefficients play in improving model performance. By capturing both the low-frequency patterns and the fine-grained, high-frequency details, the model is better equipped to make accurate and timely predictions in real-world traffic environments. The fusion of these coefficients with the raw features ensures that the model captures a broader and more nuanced set of traffic dynamics, leading to superior performance in accident anticipation tasks.

\begin{figure}
\centering
\includegraphics[width=1\textwidth,page=1]{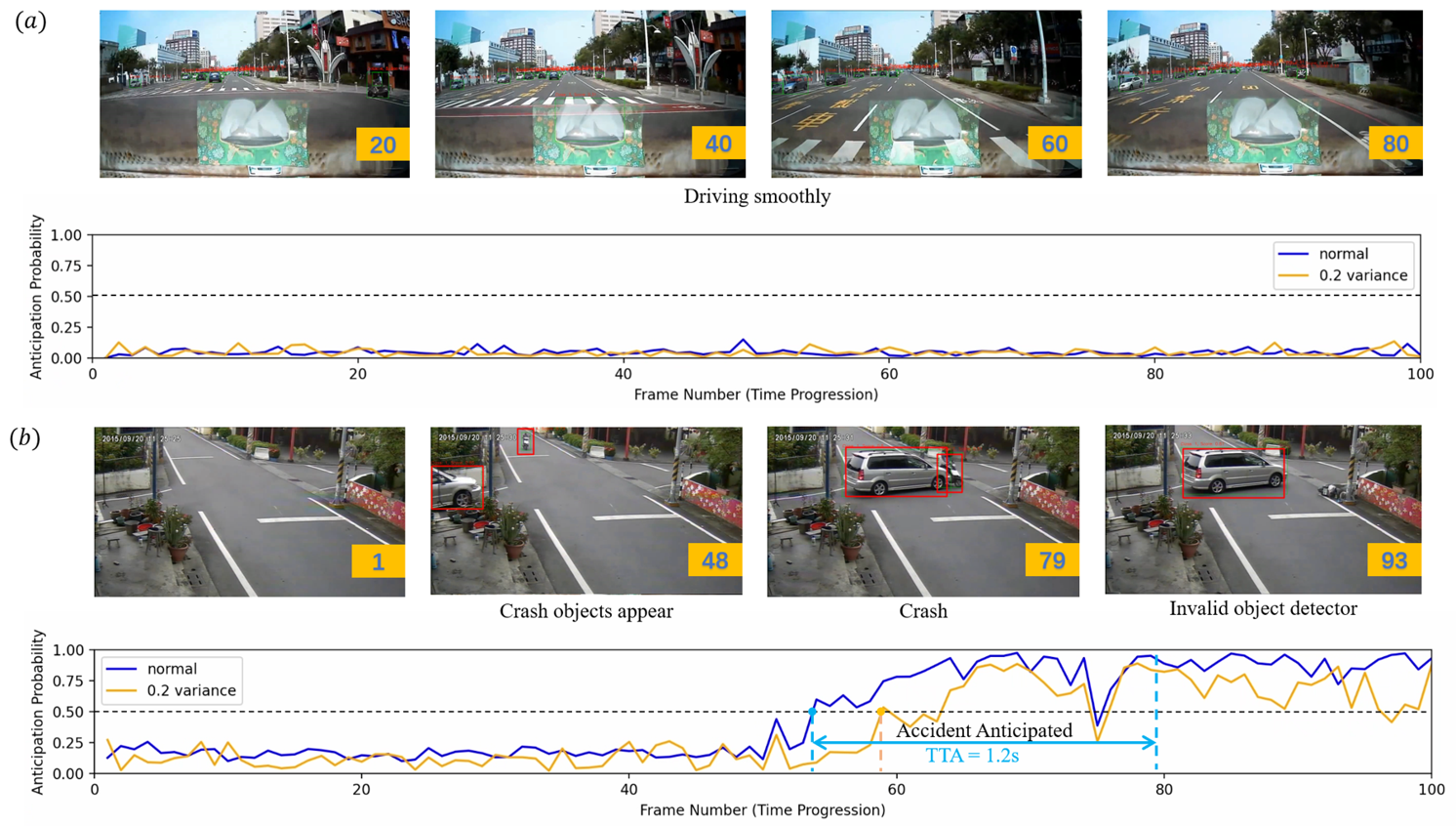}
\includegraphics[width=1\textwidth,page=2]{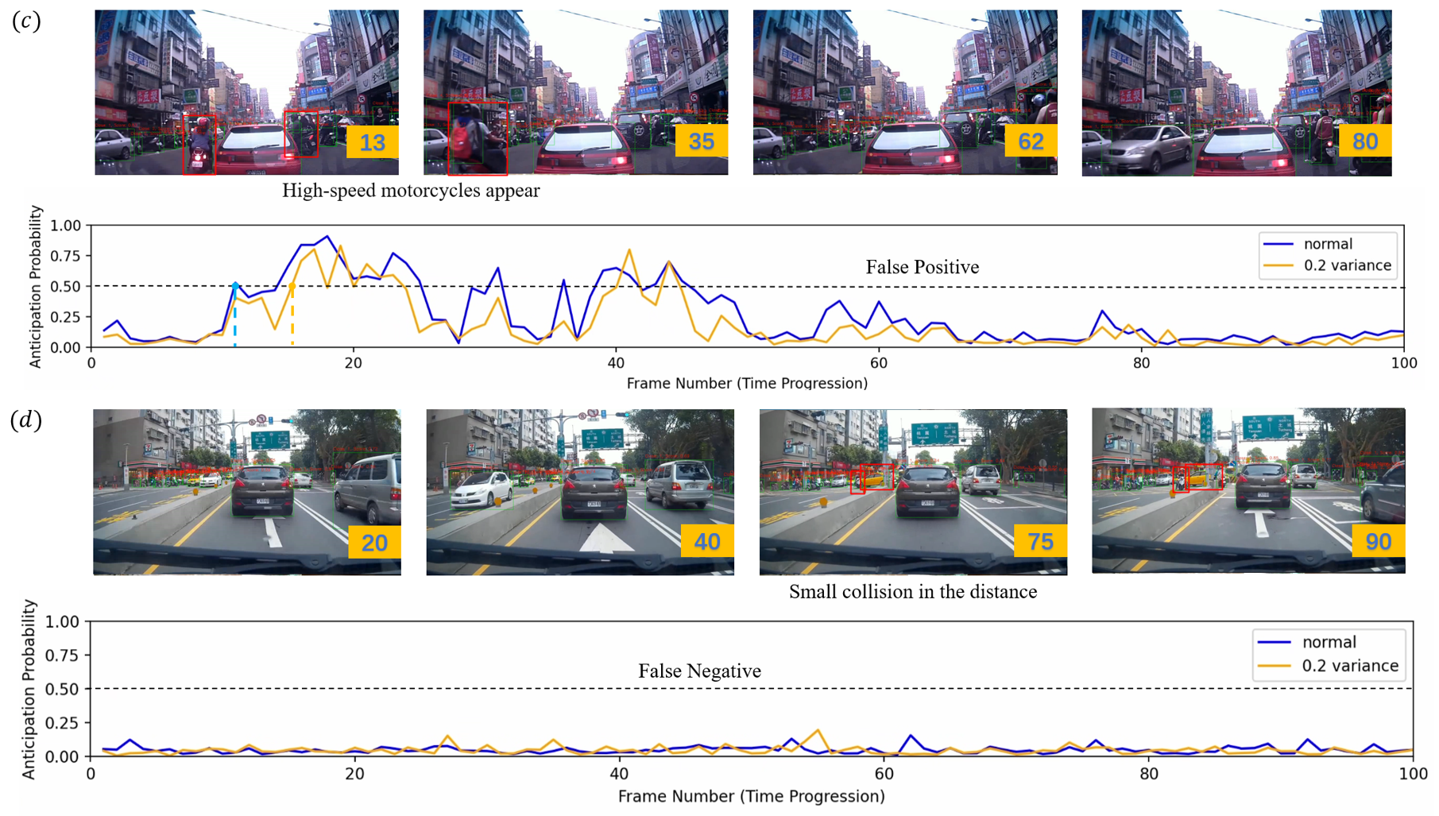} 

\caption{Visualization of ROAR's performance on the DAD dataset with 0.2 variance of Gaussian noise, with a threshold uniformly set at 0.5. Scenes (a) and (b) showcase successful accident anticipations, with (a) representing a scenario of smooth driving with no risk and (b) displaying a crash detection with a correct anticipation. Scene (c) illustrates a false-positive case where high-speed motorcycles are incorrectly anticipated as a collision, while (d) demonstrates a false-negative case, where a small collision in the distance is missed by the model. The accident anticipation probabilities are plotted over time, highlighting ROAR's ability to handle real-world challenges, including noise and dynamic traffic conditions. The model adjusts its predictions based on the context, even in situations with false alarms.}
\label{tab:Fig. 4}
\end{figure}

\subsection{Visualization of the Results}
In this subsection, we present four visual examples that demonstrate ROAR’s performance in predicting traffic accidents. Each scenario is visualized with the predicted accident probabilities plotted over time, showcasing the model’s ability to handle different real-world driving situations. Gaussian noise with a mean of 0 and a variance of 0.2 is added to simulate realistic disturbances, and the threshold for predicting accidents is set at 0.5.

\textbf{(a) Driving smoothly:} This scenario depicts normal driving with no imminent crash. The model assigns a consistently low probability for an accident as the vehicle moves through a typical traffic scene. Even with noise introduced into the data, the model maintains a low accident probability, correctly indicating no risk. This serves as a representation of a situation where the model makes accurate predictions by assigning a low risk to normal driving conditions.

\textbf{(b) Crash objects appear:} In this scenario, the model detects an accident scenario after crash objects become visible. The probability increases as the crash occurs, and the model anticipates the accident well in advance, with the Time-to-Accident (TTA) marked as 1.2 seconds. Despite the presence of noise, the model predicts the accident with high accuracy, making the appropriate adjustment in response to the high-risk elements visible in the scene.

\textbf{(c) High-speed motorcycles appear (False Positive):} In this case, the model predicts an accident when fast-moving motorcycles appear in the scene. However, no accident actually occurs. The model assigns a high probability of an accident during the presence of these motorcycles but fails to recognize that no collision is imminent. In the second half of the video, after the high-speed motorcycle is out of view, the other vehicles are running smoothly, resulting in a noticeable decrease from the high probability. This false positive is a typical challenge in high-velocity environments, where the model might misinterpret the risk based on the speed and movement of other vehicles. Despite the prediction being incorrect, the model’s sensitivity to dynamic changes in the environment reflects a cautious approach, similar to human judgment when faced with rapid-moving objects.

\textbf{(d) Small collision in the distance (False Negative):} In this scenario, a small collision occurs in the distance, which the model fails to predict. The probability remains low throughout the scene, as the model does not register the minor collision. This false negative highlights the difficulty in predicting subtle incidents that occur far from the vehicle. Although the incident may seem less significant, it is still a potential hazard. The missed prediction is reflective of the challenge in detecting distant or low-impact events, which is common in both automated and human-driven anticipation.

These examples showcase ROAR's robustness in handling diverse traffic situations, from high-speed motorcycles to subtle collisions in the distance. The model successfully anticipates accidents in many cases, but like human judgment, it occasionally faces challenges, such as false positives and false negatives. Nevertheless, the model demonstrates its potential to enhance safety by accurately identifying risks in complex, noisy driving environments.

\section{Conclusion}
This paper introduced ROAR, a robust accident anticipation model for autonomous driving, addressing challenges such as sensor noise, data imperfections, and class imbalance. By combining the Discrete Wavelet Transform (DWT), self-adaptive object-aware module, temporal attention fusion, and dynamic focal loss, ROAR outperforms state-of-the-art models in both accuracy (AP) and timeliness (mTTA) across multiple real-world datasets, including DAD, CCD, and A3D. Ablation studies demonstrated the effectiveness of each component, with DWT and dynamic focal loss significantly enhancing model performance. Despite some false positives and negatives, ROAR reliably anticipates accidents, proving its potential for real-world applications in autonomous driving.

In summary, ROAR provides a promising solution for reliable and accurate accident anticipation in complex traffic environments. Future work could explore the model’s scalability, real-time application, and its ability to generalize across different traffic conditions, further enhancing the safety of autonomous vehicles and intelligent transportation systems.

\bibliographystyle{elsarticle-num-names}
\bibliography{main}
\end{document}